\documentclass[11pt]{article}
\usepackage{eacl2017}
\usepackage{times}
\usepackage{url}
\usepackage{latexsym}

\eaclfinalcopy 

\usepackage[utf8]{inputenc}
\usepackage[T1]{fontenc}

\usepackage{multirow}

\usepackage{siunitx}
\usepackage{tikz}
\usepackage{pgfplots}

\usepackage{color}

\makeatletter
\newcommand{\@BIBLABEL}{\@emptybiblabel}
\newcommand{\@emptybiblabel}[1]{}
\makeatother

\usepackage[parfill]{parskip}
\usepackage{graphicx}               
\usepackage{amssymb}
\usepackage{amsthm}
\usepackage{mathtools}
\usepackage{enumerate}

\theoremstyle{definition}
\newtheorem*{ans*}{Answer}

\newcommand{\bmx}[0]{\begin{bmatrix}}
\newcommand{\emx}[0]{\end{bmatrix}}

\newcommand{\vect}[1]{\mathbf{#1}}

\newcommand{\matr}[1]{\mathbf{#1}}

\newcommand{\vc}[0]{\vect{c}}

\newcommand{\vh}[0]{\vect{h}}
\newcommand{\vv}[0]{\vect{v}}

\newcommand{\vz}[0]{\vect{z}}

\newcommand{\vs}[0]{\vect{s}}

\newcommand{\vr}[0]{\vect{r}}

\newcommand{\vt}[0]{\vect{t}}
\newcommand{\mW}[0]{\matr{W}}

\newcommand{\mE}[0]{\matr{E}}

\newcommand{\mU}[0]{\matr{U}}

\newcommand{\ola}{\overleftarrow}
\newcommand{\ora}{\overrightarrow}

\title{Nematus: a Toolkit for Neural Machine Translation}

\author{
Rico Sennrich$^\dag$  \quad Orhan Firat$^\star$ \quad Kyunghyun Cho$^\ddag$ \quad Alexandra Birch$^\dag$\\
{\bf Barry Haddow$^\dag$} \quad {\bf Julian Hitschler$^\P$} \quad {\bf Marcin Junczys-Dowmunt$^\dag$} \quad {\bf Samuel Läubli$^\S$}\\
 {\bf Antonio Valerio Miceli Barone$^\dag$} \quad {\bf Jozef Mokry$^\dag$} \quad {\bf Maria Nădejde$^\dag$}\\
 $^\dag$University of Edinburgh \quad $^\star$Middle East Technical University\\
 $^\ddag$New York University \quad  $^\P$Heidelberg University \quad $^\S$University of Zurich\\
}

\date{}

\begin{document}
\maketitle
\begin{abstract}
We present Nematus, a toolkit for Neural Machine Translation.
The toolkit prioritizes high translation accuracy, usability, and extensibility.
Nematus has been used to build top-performing submissions to shared translation tasks at WMT and IWSLT,
and has been used to train systems for production environments.
\end{abstract}

\section{Introduction}

Neural Machine Translation (NMT) \cite{DBLP:journals/corr/BahdanauCB14,DBLP:conf/nips/SutskeverVL14} has recently established itself as a new state-of-the art in machine translation.
We present Nematus\footnote{available at {\scriptsize \url{https://github.com/rsennrich/nematus}}}, a new toolkit for \textbf{Ne}ural \textbf{Ma}chine \textbf{T}ranslation.

Nematus has its roots in the dl4mt-tutorial.\footnote{{\scriptsize \url{https://github.com/nyu-dl/dl4mt-tutorial}}}
We found the codebase of the tutorial to be compact, simple and easy to extend, while also producing high translation quality.
These characteristics make it a good starting point for research in NMT.
Nematus has been extended to include new functionality based on recent research,
and has been used to build top-performing systems to last year's shared translation tasks at WMT \cite{sennrich-haddow-birch:2016:WMT} and IWSLT \cite{iwslt2016-uedin}.

Nematus is implemented in Python, and based on the Theano framework \cite{2016arXiv160502688short}.
It implements an attentional encoder--decoder architecture similar to \newcite{DBLP:journals/corr/BahdanauCB14}.
Our neural network architecture differs in some aspect from theirs, and we will discuss differences in more detail.
We will also describe additional functionality, aimed to enhance usability and performance, which has been implemented in Nematus.

\section{Neural Network Architecture}

Nematus implements an attentional encoder--decoder architecture similar to the one described by \newcite{DBLP:journals/corr/BahdanauCB14}, but with several implementation differences.
The main differences are as follows:

\begin{itemize}
\item We initialize the decoder hidden state with the mean of the source annotation, rather than the annotation at the last position of the encoder backward RNN.
\item We implement a novel conditional GRU with attention.
\item In the decoder, we use a feedforward hidden layer with $\tanh$ non-linearity rather than a $\text{maxout}$ before the softmax layer.
\item In both encoder and decoder word embedding layers, we do not use additional biases.
\item Compared to \textit{Look}, \textit{Generate}, \textit{Update} decoder phases in \newcite{DBLP:journals/corr/BahdanauCB14}, we implement \textit{Look}, \textit{Update}, \textit{Generate} which drastically simplifies the decoder implementation (see Table~\ref{tb:lug}).
\item Optionally, we perform recurrent Bayesian dropout \cite{2015arXiv151205287G}.
\item Instead of a single word embedding at each source position, our input representations allows multiple features (or ``factors'') at each time step, with the final embedding being the concatenation of the embeddings of each feature \cite{sennrich-haddow:2016:WMT}.
\item We allow tying of embedding matrices \cite{DBLP:journals/corr/PressW16,DBLP:journals/corr/InanKS16}.
\end{itemize}

\begin{table}[h!]
\centering
\caption{Decoder phase differences}
\label{tb:lug}
\resizebox{\columnwidth}{!}{%
\begin{tabular}{ll|ll}
\multicolumn{2}{c|}{RNNSearch \cite{DBLP:journals/corr/BahdanauCB14}} & \multicolumn{2}{c}{Nematus (DL4MT)} \\ \hline \hline
Phase       & Output - Input   & Phase        & Output - Input     \\ \hline
Look        &  $\vc_j \leftarrow \vs_{j-1}, \text{C}$ & Look         & $\vc_j \leftarrow \vs_{j-1}, y_{j-1}, \text{C}$ \\
Generate    &  $y_j \leftarrow \vs_{j-1}, y_{j-1}, \vc_j$                & Update       &   $\vs_j \leftarrow \vs_{j-1}, y_{j-1}, \vc_j$                 \\
Update      &    $\vs_j \leftarrow \vs_{j-1}, y_j, \vc_j$              & Generate     & $y_j \leftarrow \vs_j, y_{j-1}, \vc_j$
\end{tabular}%
}
\end{table}%

We will here describe some differences in more detail:

Given a source sequence $(x_1, \dots,x_{T_x})$ of length $T_x$ and a target
sequence $(y_1,\dots,y_{T_y})$ of length $T_y$, let $\vh_i$ be the annotation of the source symbol 
at position $i$, obtained by concatenating the forward and backward encoder RNN 
hidden states, $\vh_i = [ \ora{\vh}_i; \ola{\vh}_i ]$, and $\vs_j$ be the decoder hidden state at position $j$.

\paragraph{decoder initialization}

\newcite{DBLP:journals/corr/BahdanauCB14} initialize the decoder hidden state $\vs$ with the last backward encoder state.

\begin{equation*}
    \vs_0 = \tanh \left( \mW_{init}\ola{\vh}_1 \right)
\end{equation*}

with $\mW_{init}$ as trained parameters.\footnote{All the biases are omitted for simplicity.}
We use the average annotation instead:

\begin{equation*}
    \vs_0 = \tanh \left( \mW_{init}\frac{\sum_{i=1}^{T_x}\vh_i}{T_x} \right)
\end{equation*}

\paragraph{conditional GRU with attention} 

Nematus implements a novel conditional GRU with attention, cGRU$_{\text{att}}$.
A cGRU$_{\text{att}}$ uses its previous hidden state $\vs_{j-1}$, the 
whole set of source annotations $\text{C}=\lbrace\vh_1, \dots, \vh_{T_x}\rbrace$ and 
the previously decoded symbol $y_{j-1}$ in order to update its hidden state $\vs_j$, 
which is further used to decode symbol $y_j$ at position $j$,

\begin{equation*}
    \vs_j = \text{cGRU}_{\text{att}}\left(  \vs_{j-1}, y_{j-1}, \text{C}  \right)
\end{equation*}

Our conditional GRU layer with attention mechanism, 
cGRU$_{\text{att}}$, consists of three components: two 
GRU state transition blocks and an attention mechanism ATT in between.
The first transition block, $\text{GRU}_1$, combines the previous decoded symbol $y_{j-1}$ 
and previous hidden state $\vs_{j-1}$ in order to generate an intermediate 
representation $\vs^{\prime}_j$ with the following formulations:

\vspace{-10px}
{\small
\begin{align*}
    \vs_j^{\prime} = \text{GRU}_1 &  \left( y_{j-1}, \vs_{j-1}  \right) = (1 - \vz_j^{\prime}) \odot \underline{\vs}_j^{\prime} + \vz_j^{\prime} \odot \vs_{j-1},   \\
    \underline{\vs}_j^{\prime} =& ~\text{tanh} \left(   \mW^{\prime} \mE[y_{j-1}] + \vr_j^{\prime} \odot (\mU^{\prime}\vs_{j-1})  \right), \\
    \vr_j^{\prime} =& ~ \sigma \left(  \mW_r^{\prime} \mE[y_{j-1}] + \mU_r^{\prime} \vs_{j-1}  \right), \\
    \vz_j^{\prime} =& ~ \sigma \left(  \mW_z^{\prime} \mE[y_{j-1}] + \mU_z^{\prime} \vs_{j-1}  \right),
\end{align*}
}%
\noindent where $\mE$ is the target word embedding matrix, 
$\underline{\vs}_j^{\prime}$ is the proposal intermediate representation, $\vr_j^{\prime}$ 
and $\vz_j^{\prime}$ being the reset and update gate activations. In this formulation, 
$\mW^{\prime}$, $\mU^{\prime}$, $\mW_r^{\prime}$, $\mU_r^{\prime}$, 
$\mW_z^{\prime}$, $\mU_z^{\prime}$ are trained model parameters; $\sigma$ is the logistic sigmoid activation function.

The attention mechanism, ATT, inputs the entire context set C along with 
intermediate hidden state $\vs_j^{\prime}$ in order to compute the context vector 
$\vc_j$ as follows:

\begin{align*}
    \vc_j =& \text{ATT} \left(  \text{C}, \vs_j^{\prime}  \right) = \sum_i^{T_x} \alpha_{ij} \vh_i  ,   \\
     \alpha_{ij} &  = \frac{\text{exp}(e_{ij})}{\sum_{k=1}^{Tx} \text{exp}(e_{kj}) }    ,\\
    e_{ij} =& \vv_a^{\intercal} \tanh \left( \mU_a \vs_j^{\prime} + \mW_a \vh_i \right) ,    
\end{align*}

\noindent where $\alpha_{ij}$ is the normalized alignment weight between source 
symbol at position $i$ and target symbol at position $j$ and $\vv_a, \mU_a, \mW_a$ 
are the trained model parameters.

Finally, the second transition block, $\text{GRU}_2$, generates $\vs_j$, the hidden state of 
the $\text{cGRU}_{\text{att}}$, by looking at intermediate representation  
$\vs_j^{\prime}$ and context vector $\vc_j$ with the following formulations:

{\small
\begin{align*}
    \vs_j = \text{GRU}_2 & \left(  \vs_j^{\prime}, \vc_j  \right) = (1 - \vz_j) \odot \underline{\vs}_j + \vz_j \odot \vs_j^{\prime},    \\
    \underline{\vs}_j =& \tanh \left(  \mW \vc_j  + \vr_j \odot (\mU \vs_j^{\prime} )  \right) ,\\
    \vr_j =& \sigma \left( \mW_r \vc_j + \mU_r \vs_j^{\prime} \right), \\
    \vz_j =& \sigma \left( \mW_z \vc_j + \mU_z \vs_j^{\prime} \right),
\end{align*}
}%

\noindent similarly, $\underline{\vs}_j$ being the proposal hidden state, 
$\vr_j$ and $\vz_j$ being the reset and update gate activations with the 
trained model parameters $\mW, \mU, \mW_r, \mU_r, 
\mW_z, \mU_z$.

\noindent  Note that the two GRU blocks are not individually recurrent, recurrence only occurs at the level of the whole cGRU layer. This way of combining RNN blocks is similar to what is referred in the literature as \textit{deep transition} RNNs  \cite{Pascanu+et+al-ICLR2014,zilly2016recurrent}  as opposed to the more common \textit{stacked} RNNs \cite{schmidhuber1992learning,el1995hierarchical,graves2013generating}.

\paragraph{deep output}

Given $\vs_j$, $y_{j-1}$, and $\vc_j$, the output probability $p(y_j|\vs_{j},y_{j-1},\vc_j)$ is computed by a $\text{softmax}$ activation,
using an intermediate representation $\vt_j$.

{\normalsize
\begin{align*}
p(y_j|\vs_{j},&y_{j-1},\vc_j) = \text{softmax} \left( \vt_j \mW_o \right)\\
\vt_j = \tanh & \left( \vs_j \mW_{t1} + \mE[y_{j-1}] \mW_{t2} + \vc_j \mW_{t3} \right)
\end{align*}
}%

$\mW_{t1}, \mW_{t2}, \mW_{t3}, \mW_o$ are the trained model parameters.

\section{Training Algorithms}

By default, the training objective in Nematus is cross-entropy minimization on a parallel training corpus.
Training is performed via stochastic gradient descent, or one of its variants with adaptive learning rate (Adadelta \cite{zeiler2012adadelta}, RmsProp \cite{Tieleman2012}, Adam \cite{kingma2014adam}).

Additionally, Nematus supports minimum risk training (MRT) \cite{DBLP:journals/corr/ShenCHHWSL15} to optimize towards an arbitrary, sentence-level loss function.
Various MT metrics are supported as loss function, including smoothed sentence-level {\sc Bleu} \cite{chen-cherry:2014:W14-33}, METEOR \cite{denkowski:lavie:meteor-wmt:2011}, BEER \cite{stanojevic-simaan:2014:W14-33}, and any interpolation of implemented metrics.

To stabilize training, Nematus supports early stopping based on cross entropy, or an arbitrary loss function defined by the user.

\section{Usability Features}

\begin{figure}
\centering
\includegraphics[scale=0.35]{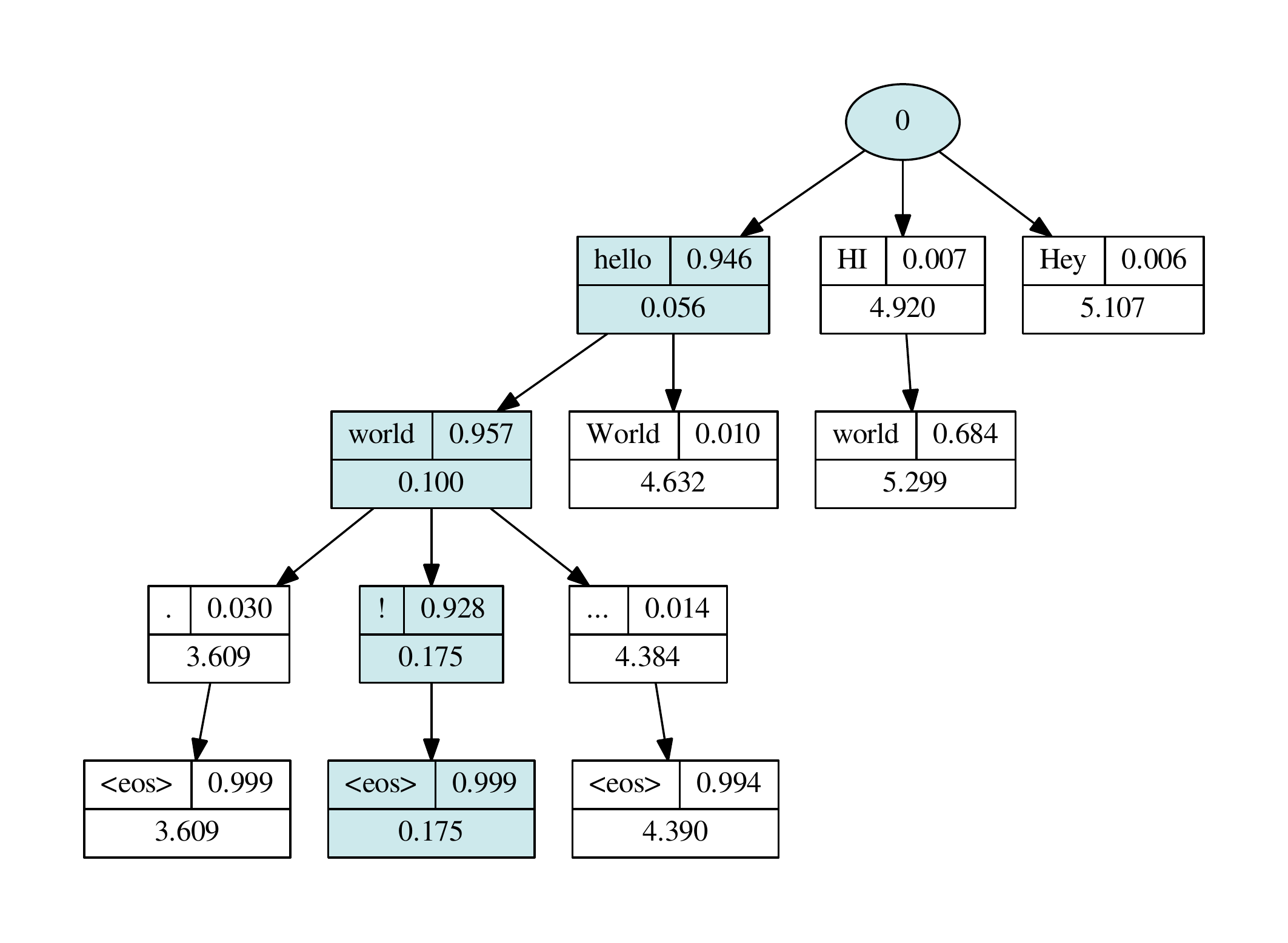}
\caption{Search graph visualisation for DE$\to$EN translation of "Hallo Welt!" with beam size 3.}
\label{fig-searchgraph}
\end{figure}

In addition to the main algorithms to train and decode with an NMT model, Nematus includes features aimed towards facilitating experimentation with the models, and their visualisation.
Various model parameters are configurable via a command-line interface, and we provide extensive documentation of options, and sample set-ups for training systems.

Nematus provides support for applying single models, as well as using multiple models in an ensemble -- the latter is possible even if the model architectures differ, as long as the output vocabulary is the same.
At each time step, the probability distribution of the ensemble is the geometric average of the individual models' probability distributions.
The toolkit includes scripts for beam search decoding, parallel corpus scoring and n-best-list rescoring.

Nematus includes utilities to visualise the attention weights for a given sentence pair, and to visualise the beam search graph.
An example of the latter is shown in Figure \ref{fig-searchgraph}.
Our demonstration will cover how to train a model using the command-line interface, and showing various functionalities of Nematus, including decoding and visualisation, with pre-trained models.\footnote{Pre-trained models for 8 translation directions are available at \scriptsize \url{http://statmt.org/rsennrich/wmt16_systems/}}

\section{Conclusion}

We have presented Nematus, a toolkit for Neural Machine Translation.
We have described implementation differences to the architecture by \newcite{DBLP:journals/corr/BahdanauCB14}; due to the empirically strong performance of Nematus, we consider these to be of wider interest.

We hope that researchers will find Nematus an accessible and well documented toolkit to support their research.
The toolkit is by no means limited to research, and has been used to train MT systems that are currently in production \cite{wipo2016}.

Nematus is available under a permissive BSD license.

\section*{Acknowledgments}

This project has received funding from the European Union's Horizon 2020 research and innovation
programme under grant agreements 645452 (QT21), 644333 (TraMOOC), 644402 (HimL) and 688139 (SUMMA).

\bibliographystyle{eacl2017}
\bibliography{bibliography.bib}

\end{document}